\documentclass[conference]{IEEEtran}
\usepackage{listings}
\usepackage{algorithm}
\usepackage[noend]{algpseudocode}
\usepackage{amsmath}
\usepackage{pdfpages}
\usepackage{siunitx}

\makeatletter
\def\BState{\State\hskip-\ALG@thistlm}
\makeatother

\ifCLASSINFOpdf
\else
\fi
\begin{document}
%
\title{An Empirical Analysis of Feature Engineering for Predictive Modeling}

\author{\IEEEauthorblockN{Jeff Heaton}
\IEEEauthorblockA{McKelvey School of Engineering\\
Washington University in St. Louis\\
St. Louis, MO 63130\\
Email: jtheaton@wustl.edu}}


%


\maketitle

\begin{abstract}
Machine learning models, such as neural networks, decision trees, random forests, and gradient boosting machines, accept a feature vector, and provide a prediction.  These models learn in a supervised fashion where we provide feature vectors mapped to the expected output.  It is common practice to engineer new features from the provided feature set.  Such engineered features will either augment or replace portions of the existing feature vector.  These engineered features are essentially calculated fields based on the values of the other features.  

Engineering such features is primarily a manual, time-consuming task.  Additionally, each type of model will respond differently to different kinds of engineered features.  This paper reports empirical research to demonstrate what kinds of engineered features are best suited to various machine learning model types.  We provide this recommendation by generating several datasets that we designed to benefit from a particular type of engineered feature.  The experiment demonstrates to what degree the machine learning model can synthesize the needed feature on its own.  If a model can synthesize a planned feature, it is not necessary to provide that feature.  The research demonstrated that the studied models do indeed perform differently with various types of engineered features. 
\end{abstract}


%
\IEEEpeerreviewmaketitle

\section{Introduction}
Feature engineering is an essential but labor-intensive component of machine learning applications \cite{bengio2013representation}.  Most machine-learning performance is heavily dependent on the representation of the feature vector. As a result, data scientists spend much of their effort designing preprocessing pipelines and data transformations \cite{bengio2013representation}. 

To utilize feature engineering, the model must preprocess its input data by adding new features based on the other features \cite{coates2011analysis}. These new features might be ratios, differences, or other mathematical transformations of existing features.  This process is similar to the equations that human analysts design. They construct new features such as body mass index (BMI), wind chill, or Triglyceride/HDL cholesterol ratio to help understand existing features' interactions.   

 Kaggle and ACM's KDD Cup have seen feature engineering play an essential part in several winning submissions.  Individuals applied feature engineering to the winning KDD Cup 2010 competition entry \cite{yu2010feature}.  Additionally, researchers won the Kaggle Algorithmic Trading Challenge with an ensemble of models and feature engineering. These individuals created these engineered features by hand.  

Technologies such as deep learning \cite{hinton2006fast} can benefit from feature engineering.  Most research into feature engineering in the deep learning space has been in image and speech recognition \cite{bengio2013representation}. Such techniques are successful in the high-dimension space of image processing and often amount to dimensionality reduction techniques \cite{hinton2006reducing} such as PCA \cite{jolliffe2002principal}  and auto-encoders \cite{olshausen1996emergence}. 

\section{Background and Prior Work}

Feature engineering grew out of the desire to transform non-normally distributed linear regression inputs.  Such a transformation can be helpful for linear regression.  The seminal work by George Box and David Cox in 1964 introduced a method for determining which of several power functions might be a useful transformation for the outcome of linear regression \cite{box1964analysis}.  This technique became known as the Box-Cox transformation.

The alternating conditional expectation (ACE) algorithm \cite{breiman1985estimating} works similarly to the Box-Cox transformation. An individual can apply a mathematical function to each component of the feature vector outcome.  However, unlike the Box-Cox transformation, ACE can guarantee optimal transformations for linear regression.

Linear regression is not the only machine-learning model that can benefit from feature engineering and other transformations.  In 1999, researchers demonstrated that feature engineering could enhance rules learning performance for text classification \cite{scott1999feature}.  Feature engineering was successfully applied to the KDD Cup 2010 competition using a variety of machine learning models.

\section{Experiment Design and Methodology}

Different machine learning model types have varying degrees of ability to synthesize various types of mathematical expressions.  If the model can learn to synthesize an engineered feature on its own, there was no reason to engineer the feature in the first place.  Demonstrating empirically a model's ability to synthesize a particular type of expression shows if engineered features of this type might be useful to that model.  To explore these relations, we created ten datasets contain the inputs and outputs that correspond to a particular type of engineered feature.  If the machine-learning model can learn to reproduce that feature with a low error, it means that that particular model could have learned that engineered feature without assistance.  

For this research, only considered regression machine learning models for this experiment.  We chose the following four machine learning models because of the relative popularity and differences in approach.

\begin{itemize}
	\item Deep Neural Networks (DANN)
	\item Gradient Boosted Machines  (GBM)
	\item Random Forests 		
	\item Support Vector Machines for Regression (SVR) 
\end{itemize}

To mitigate the stochastic nature of some of these machine learning models, each experiment was run 5 times, and the best run's outcome was used for the comparison. These experiments were conducted in the Python programming language, using the following third-party packages: Scikit-Learn \cite{pedregosa2011scikit} and TensorFlow\cite{tensorflow2015-whitepaper}.  Using this combination of packages, model types of support vector machine (SVM) \cite{vapnik1974theory}\cite{boser1992training}, deep neural network \cite{mcculloch1943logical}, random forest \cite{breiman2001random}, and gradient boosting machine (GBM) \cite{mason1999boosting} were evaluated against the following sixteen selected engineered features:

\begin{itemize}
    \item Counts
    \item Differences
    \item Distance Between Quadratic Roots
    \item Distance Formula
    \item Logarithms
    \item Max of Inputs
    \item Polynomials
    \item Power Ratio (such as BMI)
    \item Powers
    \item Ratio of a Product
    \item Rational Differences
    \item Rational Polynomials
    \item Ratios
    \item Root Distance
    \item Root of a Ratio (such as Standard Deviation)
    \item Square Roots
\end{itemize}

The techniques used to create each of these datasets are described in the following sections.  The Python source code for these experiments can be downloaded from the author's GitHub page \cite{jeffpapers} or Kaggle \cite{jeff_kaggle}.

\subsection{Counts}

The count engineered feature counts the number of elements in the feature vector that satisfies a certain condition. For example, the program might generate a count feature that counts other features above a specified threshold, such as zero.  Equation \ref{eq:countxform} defines how a count feature might be engineered.

\begin{equation} \label{eq:countxform}
y=\sum_{i=1}^n 1 \ \text{if} \ x_i > t \ \text{else} \ 0
\end{equation}

The x-vector represents the input vector of length $n$.  The resulting $y$ contains an integer equal to the number of $x$ values above the threshold ($t$). The resulting y-count was uniformly sampled from integers in the range $[1,50]$, and the program creates the corresponding input vectors for the program to generate a count dataset.  Algorithm \ref{alg:counts} demonstrates this process.

\begin{algorithm}
\caption{Generate count test dataset}\label{alg:counts}
\begin{algorithmic}[1]
\BState {\textbf{INPUT:}} The number of rows to generate $r$.
\BState {\textbf{OUTPUT:}} A dataset where $y$ contains random integers sampled from $[0, 50]$, and $x$ contains 50 columns randomly chosen to sum to $y$.
\BState {\textbf{METHOD:}}
\State {$x \gets [\text{...empty set...}] $}
\State {$y \gets [\text{...empty set...}] $}
\For{$n \gets 1 \ TO \ r$}
        \State{$v \gets \text{zeros}(50)$} \Comment{Vector of length 50}
        \State{$o \gets \text{uniform\_random\_int}(0, 50)$}\Comment{Outcome(y)}
        \State{$r \gets$ o} \Comment{remaining}
        \While{$r \ge 0$}:
            \State{$i \gets \text{uniform\_random\_int}(0, \text{len}(x) - 1)$}
            \If{$ x[i] = 0$}
                \State{$v[i] \gets 1$}
                \State{$r \gets r - 1$}
            \EndIf
          \EndWhile
        \State{$x.\text{append}(x)$}
        \State{$y.\text{append}(o)$}
\EndFor

\Return{$[x,y]$} 
\end{algorithmic}
\end{algorithm}

Several example rows of the count input vector are shown in Table \ref{tab:countxform}. The $y_1$ value simply holds the count of the number of features $x_1$ through $x_{50}$ that contain a value greater than 0.

\begin{table}[!t]
\caption{Counts Transformation}\label{tab:countxform}
\centering
\begin{tabular}{ c | c | c | c | c || c  } 
  \hline
$x_1$ & $x_2$ & $x_3$ & \ldots & $x_{50}$ & $y_1$ \\
  \hline
 1 & 0 & 1 & \ldots &  0 &  2 \\
 1 & 1 & 1 & \ldots & 1  & 12 \\
 0 & 1 & 0 & \ldots & 1  & 8 \\
 1 & 0 & 0 & \ldots & 1  & 5 \\
\hline 
\end{tabular}
\end{table}

\subsection{Differences and Ratios}

Differences and ratios are common choices for feature engineering.  To evaluate this feature type a dataset is generated with $x$ observations uniformly sampled in the real number range $[0,1]$, a single $y$ prediction is also generated that is various differences and ratios of the observations. When sampling uniform real numbers for the denominator, the range $[0.1,1]$ is used to avoid division by zero.  The equations chosen are simple difference (Equation \ref{eq:difxform}), simple ratio (Equation \ref{eq:rationxform}), power ratio (Equation \ref{eq:ratiopower}), product power ratio (Equation \ref{eq:ratioprodpower}) and ratio of a polynomial (Equation \ref{eq:ratiopoly}).

\begin{equation} \label{eq:difxform}
y=x_1 - x_2
\end{equation}

\begin{equation} \label{eq:rationxform}
y=\frac{x_1}{x_2}
\end{equation}

\begin{equation} \label{eq:ratiopower}
y=\frac{x_1}{x_2^2}
\end{equation}

\begin{equation} \label{eq:ratioprodpower}
y=\frac{x_1 x_2}{x_3^2}
\end{equation}

\begin{equation} \label{eq:ratiopoly}
y=\frac{1}{5x+8x^2}
\end{equation}

\subsection{Distance Between Quadratic Roots}

It is also useful to see how capable the four machine learning models are at synthesizing ordinary mathematical equations.  We generate the final synthesized feature from a distance between the roots of a quadratic equation.  The distance between roots of a quadratic equation can easily be calculated by taking the difference of the two outputs of the quadratic formula, as given in Equation \ref{eq:quad}, in its unsimplified form.

\begin{equation} \label{eq:quad}
y=\left| \frac{-b+\sqrt{b^2-4ac}}{2a}- \frac{-b-\sqrt{b^2-4ac}}{2a} \right|
\end{equation}

The dataset for the transformation represented by Equation \ref{eq:quad} is generated by uniformly sampling $x$ values from the real number range $[-10,10]$.  We discard any invalid results.

\subsection{Distance Formula}

The distance formula contains a ratio inside a radical, and is shown in Equation \ref{eq:dist}. The input are for $x$ values uniformly sampled from the range $[0, 10]$, and the outcome is the Euclidean distance between $(x_1, x_2)$ and $(x_3, x_4)$.

\begin{equation} \label{eq:dist}
y=\sqrt{(x_1 - x_2)^2 + (x_3 - x_4)^2} 
\end{equation}

\subsection{Logarithms and Power Functions}

Statisticians have long used logarithms and power functions to transform the inputs to linear regression \cite{box1964analysis}.  Researchers have shown the usefulness of these functions for transformation for other model types, such as neural networks \cite{balkin2000automatic}.  The log and power transforms used in this paper are of the type shown in Equations \ref{eq:logxform},\ref{eq:secpowerxform}, and \ref{eq:sqrtxform}.

\begin{equation} \label{eq:logxform}
y=\log(x)
\end{equation}

\begin{equation} \label{eq:secpowerxform}
y=x^2
\end{equation}

\begin{equation} \label{eq:sqrtxform}
y=x^{\frac{1}{2}}
\end{equation}

This paper investigates using the natural log function, the second power, and the square root.  For both the log and root transform, random $x$ values were uniformly sampled in the real number range $[1,100]$.  For the second power transformation, the $x$ values were uniformly sampled in the real number range $[1, 10]$. A single $x_1$ observation is used to generate a single $y_1$ observation.  The $x_1$ values are simply random numbers that produce the expected $y_1$ values by applying the logarithm function.

\subsection{Max of Inputs}

Ten random inputs are generated for the observations ($x_1 - x_10$).  These random inputs are sampled uniformly in the range $[1, 100]$.  The outcome is the maximum of the observations.  Equation \ref{eq:maxinput} shows how this research calculates the max of inputs feature.

\begin{equation} \label{eq:maxinput}
y=\max{(x_1 ... x_{10})}
\end{equation}

\subsection{Polynomials}

Engineered features might take the form of polynomials.  This paper investigated the machine learning models' ability to synthesize features that follow the polynomial given by Equation \ref{eq:polyxform}.

\begin{equation} \label{eq:polyxform}
y=1+5x+8x^2
\end{equation}

An equation such as this shows the models' ability to synthesize features that contain several multiplications and an exponent.  The data set was generated by uniformly sampling $x$ from real numbers in the range $[0,2)$.  The $y_1$ value is simply calculated based on $x_1$ as input to Equation \ref{eq:polyxform}.

\subsection{Rational Differences and Polynomials}

Useful features might also come from combinations of rational equations of polynomials.  Equations \ref{eq:ratdifxform} \& \ref{eq:ratpolyxform} show the types of rational combinations of differences and polynomials tested by this paper. We also examine a ratio power equation, similar to the body mass index (BMI) calculation, shown in Equations \ref{eq:bmi}.

\begin{equation} \label{eq:ratdifxform}
y=\frac{x_1-x_2}{x_3-x_4 }
\end{equation}

\begin{equation} \label{eq:ratpolyxform}
y=\frac{1}{5x+8x^2} 
\end{equation}

\begin{equation} \label{eq:bmi}
y=\frac{x_1}{x_2^2} 
\end{equation}

To generate a dataset containing rational differences (Equation \ref{eq:ratdifxform}), four observations are uniformly sampled from real numbers of the range $[1,10]$.  Generating a dataset of rational polynomials, a single observation is uniformly sampled from real numbers of the range $[1,10]$.

\section{Results Analysis}

To evaluate the effectiveness of the four model types over the sixteen different datasets we must account for the differences in ranges of the $y$ values.  As Table \ref{table:results-1} demonstrates, the maximum, minimum, mean, and standard deviation of the datasets varied considerably. Because an error metric, such as root mean square error (RMSE) is in the same units as its corresponding $y$ values, some means of normalization is needed.  To allow comparison across datasets, and provide this normalization, we made use of the normalized root-mean-square deviation (NRMSD) error metric shown in Equation \ref{eq:nrmsd}. We capped all NRMSD values at 1.5; we considered values higher than 1.5 to have failed to synthesize the feature.

\begin{equation} \label{eq:nrmsd}
\operatorname{NRMSD}=\frac{1}{\sigma}\sqrt{\frac{\sum_{t=1}^T (\hat y_t - y_t)^2}{T}}
\end{equation}

The results obtained by the experiments performed in this paper clearly indicate that some model types perform much better with certain classes of engineered features than other model types.  The simple transformations that only involved a single feature were all easily learned by all four models.  This included the log, polynomial, power, and root.  However, none of the models were able to successfully learn the ratio difference feature.  Table \ref{table:results-2} provides the scores for each equation type and model.  The model specific results from this experiment are summarized in the following sections.

\begin{table}[!t]
\caption{Dataset observation (y) statistics and ranges}\label{table:results-1}
\centering
\begin{tabular}{|l|l|l|l|l|}
\hline
\textbf{name} & \textbf{min} & \textbf{max} & \textbf{mean} & \textbf{std} \\
\hline
bmi     & 6.25      & 87.77   & 37.22    & 18.22  \\
counts  & 0.0                    & 49.0                & 24.60     & 14.47  \\
dev     & 7.56      & 41.70   & 27.01    & 4.57   \\
diff    & -0.99    & 0.98   & -0.007 & 0.40  \\
dist    & 0.07     & 11.55  & 4.71     & 2.25   \\
log     & 0.00   & 4.60   & 3.66     & 0.87  \\
max     & 37.89      & 99.99   & 90.87     & 8.38    \\
poly    & 1.00     & 42.99   & 16.79    & 12.37  \\
pow     & 1.00     & 99.99   & 37.29    & 29.25   \\
quad    & 0.0                    & 11.66  & 1.59    & 2.03    \\
r\_diff & -24,976.28       & 4,205.68     & -2.83   & 259.28  \\
r\_poly & 0.001   & 0.07 & 0.009  & 0.01 \\
ratio   & 2.36e-05 & 93.92   & 2.33     & 5.32    \\
rel     & 0.015   & 88.38   & 3.15    & 6.63  \\
sqrt    & 1.00     & 9.99   & 6.75    & 2.28   \\
sum     & 16.82     & 81.38   & 49.95     & 9.16    \\
\hline
\end{tabular}
\end{table}

\begin{table}[!t]
\caption{Model Scores for Datasets}\label{table:results-2}
\centering
\begin{tabular}{|l|l|l|l|l|}
\hline
\textbf{Feature} & \textbf{Score SVM} & \textbf{Score RF} & \textbf{Score GBM} & \textbf{Score Neural} \\
\hline
bmi      & 0.00 & 0.00 & 0.00 & 0.03 \\
counts   & 0.00 & 0.08 & 0.10 & 0.00 \\
dev      & 0.03 & 0.45 & 0.32 & 0.05 \\
diff     & 0.07 & 0.00 & 0.01 & 0.00 \\
dist     & 0.03 & 0.17 & 0.12 & 0.01 \\
log      & 0.07 & 0.00 & 0.00 & 0.00 \\
max      & 0.36 & 0.06 & 0.12 & 0.07 \\
poly     & 0.00 & 0.00 & 0.00 & 0.00 \\
pow      & 0.00 & 0.00 & 0.00 & 0.00 \\
quad     & 0.56 & 0.04 & 0.04 & 0.03 \\
r\_diff  & 0.99  & 1.39 & 1.5 & 1.00 \\
r\_poly  & 1.5 & 0.00 & 0.00 & 0.02 \\
ratio    & 0.28 & 0.03 & 0.05 & 0.03 \\
rel      & 0.07 & 0.05 & 0.09 & 0.02 \\
sqrt     & 0.03 & 0.00 & 0.00 & 0.00\\
sum      & 0.00 & 0.33 & 0.28 & 0.00 \\
\hline
\end{tabular}
\end{table}

\subsection{Neural Network Results}

For each engineered feature experiment, create an ADAM \cite{kingma2014adam} trained deep neural network.  We made use of a learning rate of 0.001, $\beta_1$ of 0.9, $\beta_2$ of 0.999, and $\epsilon$ of $1 \times 10^{-7}$, the default training hyperparameters for Keras ADAM.

\begin{figure}[!t]
   \centering      
   \includegraphics[width=8cm]{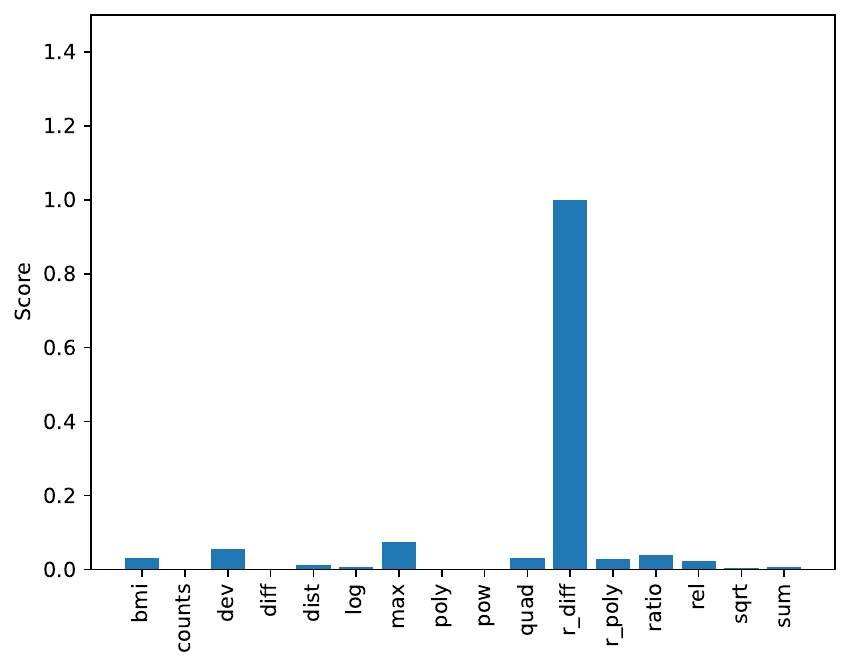}      
 \caption{Deep Neural Network Engineered Features}
 \label{fig:error_dann}
\end{figure}

The deep neural network contained the number of input neurons equal to the number of inputs needed to test that engineered feature type.  Likewise, a single output neuron provided the value generated by the specified engineered feature.  When viewed from the input to the output layer, there are five hidden layers, containing 400, 200, 100, 50, and 25 neurons, respectively.  Each hidden layer makes use of a rectifier transfer function \cite{glorot2011deep}, making each hidden neuron a rectified linear unit (ReLU). We provide the results of these deep neural network engineered feature experiments in Figure \ref{fig:error_dann}.

The deep neural network performed well on all equation types except the ratio of differences.  The neural network also performed consistently better on the remaining equation types than the other three models. An examination of the calculations performed by a neural network will provide some insight into this performance. A single-layer neural network is essentially a weighted sum of the input vector transformed by a transfer function, as shown in Equation \ref{eq:neuron}.

\begin{equation} \label{eq:neuron}
f(x,w,b)= \phi \left( \sum_{n} (w_i  x_i) + b \right)
\end{equation}

The vector $x$ represents the input vector, the vector $w$ represents the weights, and the scalar variable $b$ represents the bias.  The symbol $\phi$ represents the transfer function.  This paper's experiments used the rectifier transfer function \cite{glorot2011deep} for hidden neurons and a simple identity linear function for output neurons.  The weights and biases are adjusted as the neural network is trained.  A deep neural network contains many layers of these neurons, where each layer can form the input (represented by $x$) into the next layer.  This fact allows the neural network to be adjusted to perform many mathematical operations and explain some of the results shown in Figure \ref{fig:error_dann}.  The neural network can easily add, sum, and multiply.  This fact made the counts, diff, power, and rational polynomial engineered features all relatively easy to synthesize by using layers of Equation \ref{eq:neuron}.  

\subsection{Support Vector Machine Results}

\begin{figure}[!t]
   \centering      
   \includegraphics[width=8cm]{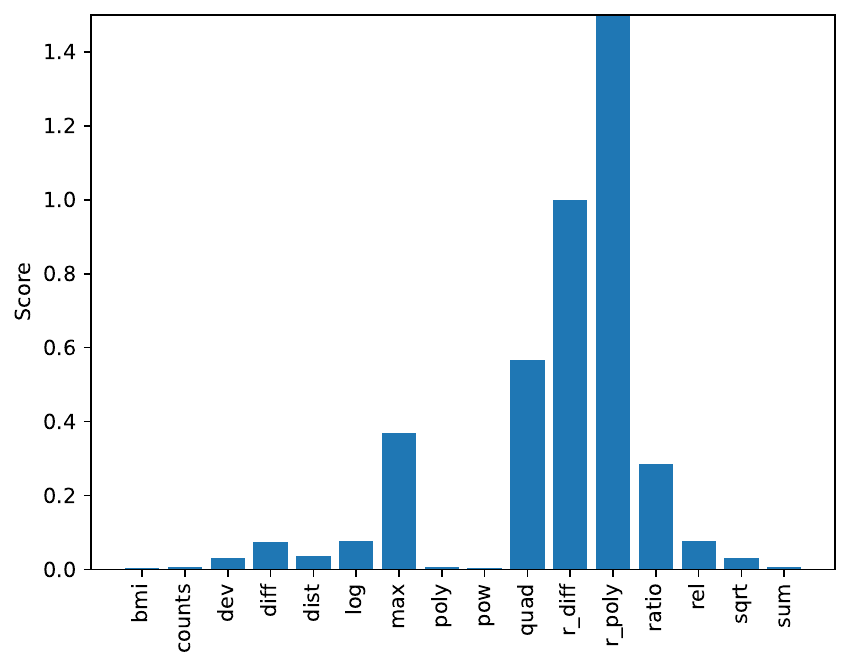}      
 \caption{SVM Engineered Features}
 \label{fig:error_svr}
\end{figure}

The two primary hyper-parameters of an SVM are $C$ and $\gamma$.  It is customary to perform a grid search to find an optimal combination of $C$ and $\gamma$ \cite{hsu2003practical}.  We tried 3 $C$ values of 0.001, 1, and 100, combined with the 3 $\gamma$ values of 0.1, 1, and 10. This selection resulted in 9 different SVMs to evaluate. The experiment results are from the best combination of $C$ and $\gamma$ for each feature type.  A third hyper-parameter specifies the type of kernel that the SVM uses, which is a Gaussian kernel.  Because support vector machines benefit from their input feature vectors normalized to a specific range \cite{hsu2003practical}, we normalized all SVM input to [0,1].  This required normalization step for the SVM does add additional calculations to the feature investigated.  Therefore, the SVM results are not as pure of a feature engineering experiment as the other models. We provide the results of the SVM engineered features in Figure \ref{fig:error_svr}.

The support vector machine found the max, quadratic, ratio of differences, a polynomial ratio, and a ratio all difficult to synthesize.  All other feature experiments were within a low NRMSD level. 
Smola and Vapnik extended the original support vector machine to include regression; we call the resulting algorithm a support vector regression (SVR) \cite{smola1997support}. A full discussion of how an SVR is fitted and calculated is beyond the scope of this paper.  However, for this paper's research, the primary concern is how an SVR calculates its final output.  This calculation can help determine the transformations that an SVR can synthesize.  The final output for an SVR is given by the decision function, shown in Equation \ref{eq:svr}.

\begin{equation} \label{eq:svr}
y = \sum_{i=1}^{n} (\alpha_i - \alpha_i^{*})K(x_i,x)+\rho
\end{equation}

The vector $x$ represents the input vector; the difference between the two alphas is called the SVR's coefficient.  The weights of the neural network are somewhat analogous to the coefficients of an SVR.  The function $K$ represents a kernel function that introduces non-linearity.  This paper used a radial basis function (RBF) kernel based on the Gaussian function.  The variable $\rho$ represents the SVR intercept, which is somewhat analogous to the bias of a neural network.  

Like the neural network, the SVR can perform multiplications and summations.  Though there are many differences between a neural network and SVR, the final calculations share many similarities.  

\subsection{Random Forest Results}

\begin{figure}[!t]
   \centering      
   \includegraphics[width=8cm]{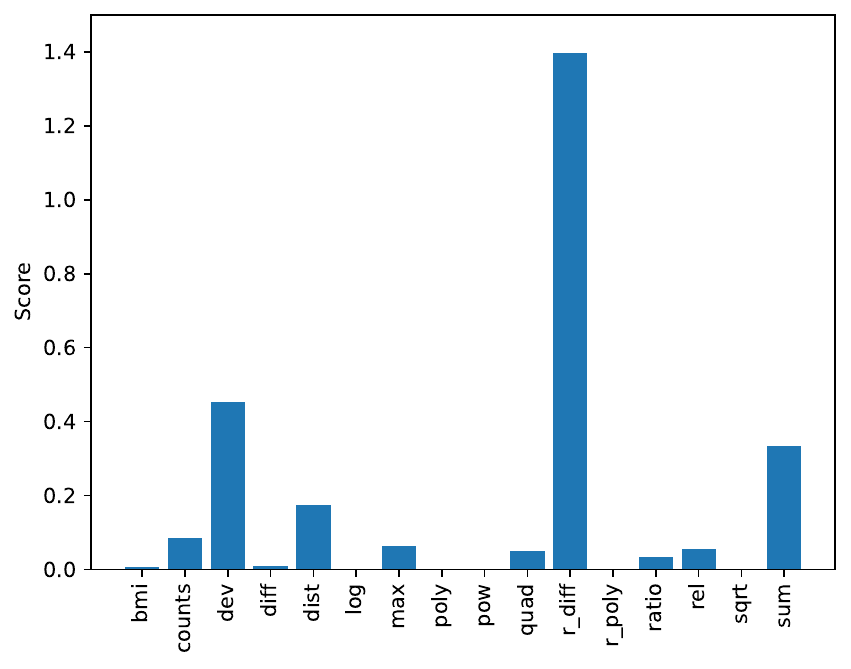}      
 \caption{Random Forest Engineered Features }
 \label{fig:error_rf}
\end{figure}

Random forests are an ensemble model made up of decision trees.  We randomly sampled the training data to produce a forest of trees that together will usually outperform the individual trees.  The random forests used in this paper all use 100 classifier trees.  This tree count is a hyper-parameter for the random forest algorithm.  We show the result of the random forest model's attempt to synthesize the engineered features in Figure \ref{fig:error_rf}.

The random forest model had the most difficulty with the standard deviation, a ratio of differences, and sum.

\subsection{Gradient Boosted Machine}

\begin{figure}[!t]
   \centering      
   \includegraphics[width=8cm]{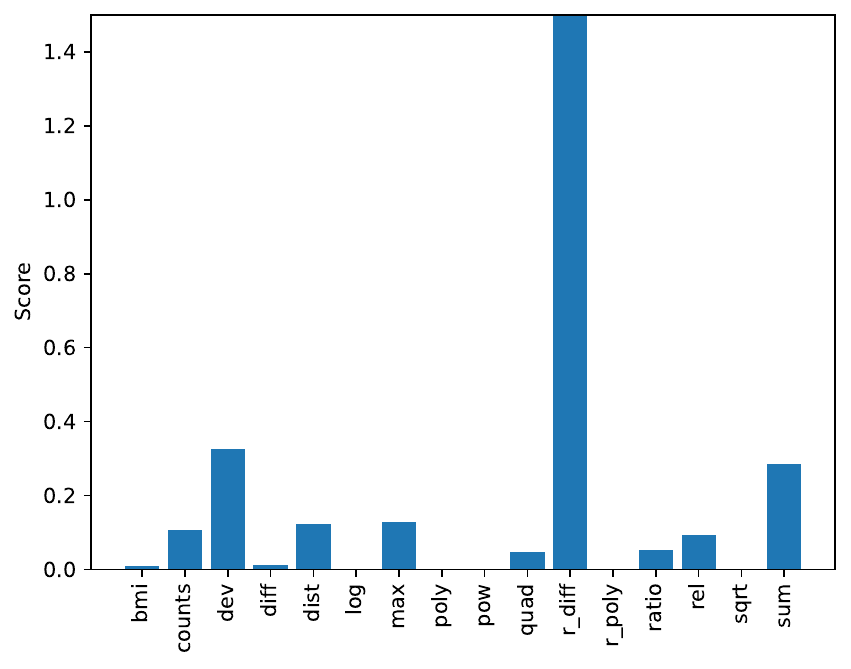}      
 \caption{Figure 4: GBM Engineered Features }
 \label{fig:error_gbm}
\end{figure}

The gradient boosted machine (GBM) model operates very similarly to random forests.  However, the GBM algorithm uses the gradient of the training objective to produce optimal combinations of the trees.  This additional optimization sometimes gives GBM a performance advantage over random forests.  The gradient boosting machines used in this paper all used the same hyper-parameters.  The maximum depth was ten levels, the number of estimators was 100, and the learning rate was 0.05.  We provide the results of the GBM engineered features in Figure \ref{fig:error_gbm}.

Like the random forest model, the gradient boosted machine had the most difficulty with the standard deviation, the ratio of differences, and sum.

\section{Conclusion \& Further Research}

Figures 1-4 clearly illustrate that machine learning models such as neural networks, support vector machines, random forests, and gradient boosting machines benefit from a different set of synthesized features.  Neural networks and support vector machines generally benefit from the same types of engineered features; similarly, random forests and gradient boosting machines also typically benefit from the same set of engineered features.  The results of this research allow us to make recommendations for both the types of features to use for a particular machine learning model type and the types of models that will work well with each other in an ensemble.

 Based on the experiments performed in this research, the type of machine learning model used has a great deal of influence on the types of engineered features to consider.  Engineered features based on a ratio of differences were not synthesized well by any of the models explored in this paper.  Because these ratios of difference might be useful to a wide array of models, all models explored here might benefit from engineered features based on ratios with differences.    
 
The research performed by this paper also empirically demonstrates one of the reasons why ensembles of models typically perform better than individual models.  Because neural networks and support vector machines can synthesize different features than random forests and gradient boosting machines, ensembles made up of a model from each of these two groups might perform very well. A neural network or support vector machine might ensemble well with a random forest or gradient boosting machine.  

We did not spend significant time tuning the models for each of the datasets. Instead, we made reasonably generic choices for the hyper-parameters chosen for the models.  Results for individual models and datasets might have shown some improvement for additional time spent tuning the hyper-parameters. 

Future research will focus on exploring other engineered features with a wider set of machine learning models.  Engineered features that are made up of multiple input features seem a logical focus.  

This paper examined 16 different engineered features for four popular machine learning model types.  Further research is needed to understand what features might be useful for other machine learning models.  Such research could help guide the creation of ensembles that use a variety of machine learning model types.  We might also examine additional types of engineered features.  It would be useful to see how more complex classes of features affect machine learning models' performance.

\bibliographystyle{IEEEtran}
\bibliography{jheaton_southeastcon2016_features}

\end{document}